\documentclass{article}
\usepackage{spconf,amsmath,graphicx,subfigure}
\usepackage{hyperref}
\usepackage{units}

\hypersetup{
	colorlinks   = true,    
	urlcolor     = blue,    
	linkcolor    = red,     
	citecolor    = green    
}

\title{Drone Shadow Tracking}

\name{Xiaoyan Zou$^{1}$ \qquad Ruofan Zhou \qquad Majed El Helou \qquad Sabine S\"{u}sstrunk
\thanks{$^{1}$ Work done in majority during a semester project in IVRL.}}
\address{School of Computer and Communication Sciences, EPFL, Switzerland}
\begin{document}
\maketitle
\begin{abstract}
Aerial videos taken by a drone not too far above the surface may contain the drone's shadow projected on the scene. This deteriorates the aesthetic quality of videos. With the presence of other shadows, shadow removal cannot be directly applied, and the shadow of the drone must be tracked.

Tracking a drone's shadow in a video is, however, challenging. The varying size, shape, change of orientation and drone altitude pose difficulties. The shadow can also easily disappear over dark areas. However, a shadow has specific properties that can be leveraged, besides its geometric shape. In this paper, we incorporate knowledge of the shadow's physical properties, in the form of shadow detection masks, into a correlation-based tracking algorithm. We capture a test set of aerial videos taken with different settings and compare our results to those of a state-of-the-art tracking algorithm.
\end{abstract}

\let\thefootnote\relax\footnotetext{https://github.com/IVRL/Drone-Shadow-Tracking}

\section{Introduction} \label{sec:intro}
Drones have gained in popularity in recent years. They are used in object deliveries, search and rescue operations, or for entertainment in racing competitions. One popular use of drones is the capture of aerial videos. However, the presence of the drone's shadow moving in the scene is an undesirable artifact in the aerial drone videos. 

\begin{figure}[t!]
	\centering
	\subfigure[Input frame]{
		\includegraphics[width=0.47\linewidth]{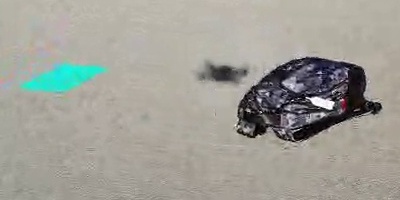}
	}
	\subfigure[MOSSE tracking~\cite{mosse}]{
		\includegraphics[width=0.47\linewidth]{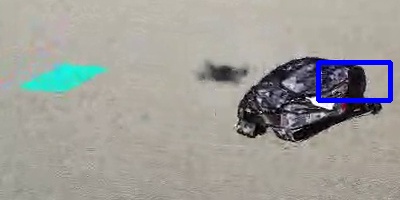}
	}
	\subfigure[Our tracking]{
		\includegraphics[width=0.47\linewidth]{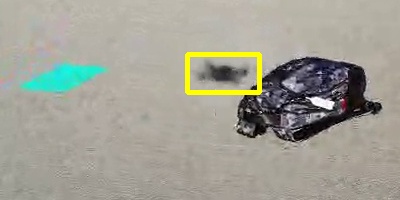}
	}
	\subfigure[Shadow-free frame]{
		\includegraphics[width=0.47\linewidth]{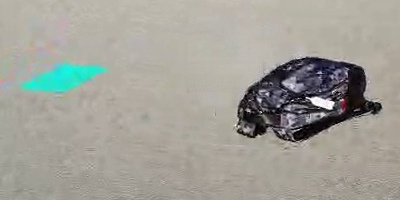}
	}
	\caption{(a) Input frame taken from an aerial drone video. (b) Drone shadow tracking result~\cite{mosse}. (c) Drone shadow tracking result, obtained with our proposed algorithm. (d) Drone shadow removed by inpainting~\cite{inpaintingyu2018}.}
	\label{fig:intro}
\end{figure}

Shadow removal cannot be applied directly on the video frames since other shadows, which need to be retained, may be naturally present in the scene. Therefore, the shadow of the drone must be tracked before a video can be processed. 
However, tracking a drone's shadow, even with state-of-the-art object tracking, is often prone to failure. A commercial drone is small, its shadow contains holes, and it flies relatively high making its shadow even smaller. Flare may make the shadow appear blurry. A drone changes orientation, as it rotates in the air to change direction, making its shadow change shape. The shadow can also change shape when the drone moves up or down, or based on the geometry of objects it is projected on. 

The biggest issue remains, however, that the shadow easily disappears over dark areas or other shadow areas in the filmed scene. One key factor that we can leverage is the fact that the object we are tracking has specific physical properties irrespective of its shape, because it is a shadow. 

We combine physical and geometrical information in correlation-based tracking to avoid losing the shadow that is being tracked. We present an approach tailored to shadow tracking, and evaluate it on tracking the shadow of a flying drone. We propose an algorithm for predicting when the shadow is likely lost by analyzing shadow detection masks, and leverage it to improve the overall tracking performance. This shadow information is then combined with correlation results in our proposed shadow fusion tracking. Our method builds on state-of-the-art object tracking through correlation kernel filtering.

\section{Related Work} \label{sec:related_work}
Object tracking methods can be classified into point tracking, kernel tracking, silhouette tracking~\cite{survey,categories2}, and the recent, computationally more expensive, deep learning approaches. A thorough tracking survey is presented in~\cite{survey}.
Point tracking methods generally leverage filters to extract object features~\cite{point_track}, used by the tracking algorithm.
Kernel tracking methods~\cite{kernel_track} are based on the appearance of the tracked objects and rely largely on direct matching. 
Silhouette tracking methods build a model for the tracked object~\cite{silhouette}, using a set of previous frames in the tracking video.

State-of-the-art visual tracking generally relies on discriminative correlation filters~\cite{kcf}. One such example is the Minimum Output Sum of Squared Error (MOSSE) method~\cite{mosse}.
Explicit geometric feature extraction from a drone shadow is complicated by the characteristics of shadows. Indeed, they change both size and shape following the drone's positioning, they are small and blurry. Moreover, they present very few edges, and are affected by the surface on which they are projected. Therefore, we rely on a discriminative correlation filter strategy. It avoids explicit feature extraction and is a fast enough candidate for real-time operation, with no need for GPUs. A review of correlation filter based tracking can be found in~\cite{mosse}. Our method also leverages correlation-based search in the tracking procedure. We compare it, on the task of shadow tracking, to the state-of-the-art deep learning tracking~\cite{nips} and correlation-based tracking~\cite{mosse}.

There exists a multitude of shadow detection (and removal) algorithms~\cite{survey_removal}. They leverage pixel intensity properties~\cite{tian,text,nir,nirour,refle,ma} and texture features~\cite{refle,guo,shluo,texture}. Image luminance~\cite{nir,nirour,refle,guo} or reflectance properties~\cite{tian,shluo} are used to find shadow boundaries. Illumination field clustering~\cite{shluo} can also segment shadows in an image. We use an off-the-shelf shadow detection method. High-quality shadow detection (and removal) can also be achieved with near-infrared (NIR) images~\cite{nir,nirour}. NIR-equipped drones and any other future improvements in shadow detection can be directly integrated to improve our proposed algorithm's robustness.

\section{Method}\label{sec:method}
We present the building blocks of our shadow tracking method. Our approach combines state-of-the-art kernel-based tracking (Sec.~\ref{sec:track}) and shadow detection (Sec.~\ref{sec:shadow}). 

\subsection{Kernel-based tracking}\label{sec:track}
The proposed algorithm works with any correlation filter method. In this paper, we use the MOSSE method proposed by Bolme~\emph{et al.}~\cite{mosse}. Its tracking is based on Fourier-domain correlation filters to leverage the computational benefits of the Fast Fourier Transform~\cite{fourier}.  
MOSSE runs in two phases, the filter initialization phase and the object tracking phase.

In order to initialize its kernel, the algorithm requires user input. The user draws a bounding box around the tracking target, which in our case is the drone's shadow. The patch containing the target is called $F$.
The filter $H$ is calculated by minimizing the sum of squared error in the Fourier domain
\begin{equation}\label{eq:minH}
    H_i = \textrm{arg} \min_H  \displaystyle\sum_{i} |F_{i}\odot H^{*} - G_{i}|^{2},
\end{equation}
where $*$ indicates the complex conjugate, and $G_{i}$ is the correlation map. A closed-form solution is given in~\cite{mosse} by
\begin{equation}\label{eq:filterH}
    H^{*}_i = \frac{ \sum_{i} G_{i} \odot F_{i}^{*}}{ \sum_{i} F_{i} \odot F_{i}^{*}}.
\end{equation}
In the object tracking phase, the filter $H_i$ and the patch $F_{i+1}$ from the next frame are multiplied in the Fourier domain and yield $G_{i+1}$, which is the correlation response for the next frame. 
The location of the maximal correlation point in a limited area in $G_{i+1}$ (delimited by the previous bounding box), is taken as the center of the drone's new position in the next frame. Iteratively, the algorithm predicts the shadow's positions throughout the entire video. As we discuss in more detail in Sec.~\ref{sec:lm}, our proposed algorithm adaptively extends the heat map search area and fuses the aforementioned correlation map with shadow detection results. 

\subsection{Shadow detection}\label{sec:shadow}
Shadow detection is leveraged multiple times in our proposed shadow tracking algorithm. It is a key component in the failure prediction algorithm, which predicts when the kernel-based tracking is prone to fail in tracking the shadow. It is also fused with correlation maps to improve shadow tracking. 

In this paper, we use the adaptive thresholding method proposed by Bradley et al.~\cite{shadow}, which extends the established technique of~\cite{wellner} to scenarios with illumination changes. The algorithm traverses the image in grayscale through a window of size $16 \times 16$, computing the mean intensity, with an integral technique, of every windowed region. In the last step, the mean value of the region is compared to neighboring pixels to determine whether they are shadow pixels or not, to produce a binary shadow mask.

Our fusion approach is independent of the specific shadow detection method used. Any shadow detector can be integrated into the tracking algorithm. Improvements in shadow detection quality increase the tracking robustness.

\subsection{Tracking-failure prediction}\label{sec:failmosse}
We describe in this section our failure prediction algorithm that classifies  a given frame as having a tracking error or not. 
The failure prediction takes as input the shadow detection result of the current frame, and, from the previous frame, the position of the bounding box and the area of the drone's shadow. 

We reduce the noise in the shadow detection result (Sec.~\ref{sec:shadow}) through erosion and dilation with a kernel of size $3\times3$.
The algorithm computes object contours on the resulting binary shadow mask. 
The drone's shadow area at frame $i+1$ ($A_{i+1}$) is inferred from the computed contour that is taken as the contour closest to the center of the tracking bounding box. This is the reason the shadow detection result is filtered to remove noise, to avoid having noise contours being closest to the center. 
When the absolute incremental change in area is large (defined empirically as $A_{i+1} \geq 2.5\times A_i$), the method predicts failure. This is often because a large and sudden change in shadow area indicates that the drone's shadow is being mistaken with a different object by the tracking algorithm. 
When failure is not detected, regular tracking takes place. And when failure is predicted, the shadow fusion algorithm is applied.

\begin{figure}[t!]
	\centering
	\includegraphics[width=\linewidth, trim={5 0 10 0}, clip]{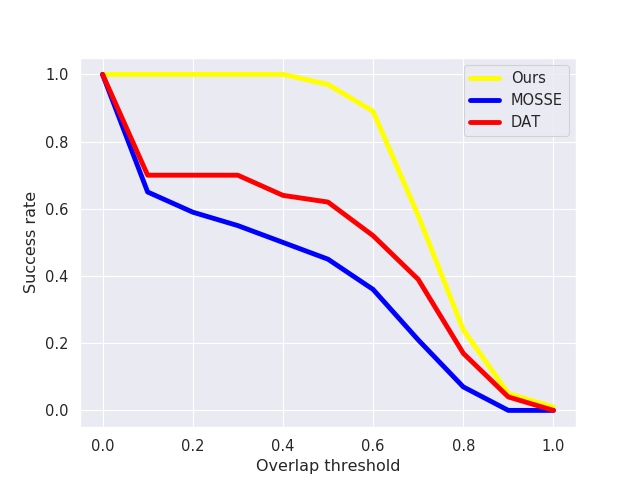}
	\caption{Average performance of MOSSE~\cite{mosse}, DAT~\cite{nips}, and our approach, over our annotated video test set.}
	\label{fig:IoU}
\end{figure}

\subsection{Shadow fusion tracking}\label{sec:lm}
\begin{figure}[t!]
	\centering
	\subfigure{
		\includegraphics[width=0.47\linewidth, trim={5 0 10 0}, clip]{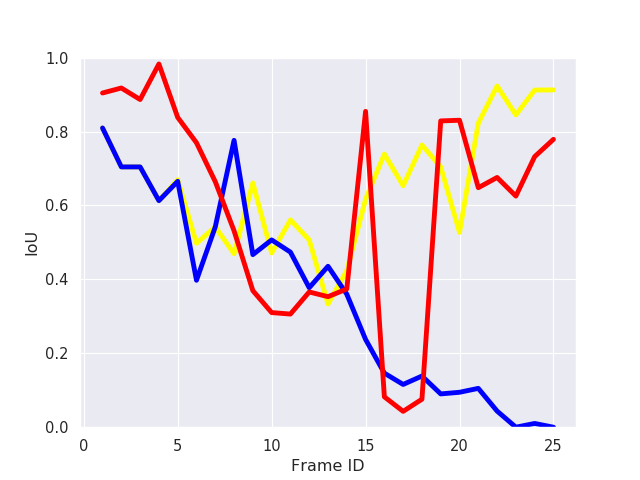}
	}
	\subfigure{
		\includegraphics[width=0.47\linewidth, trim={5 0 10 0}, clip]{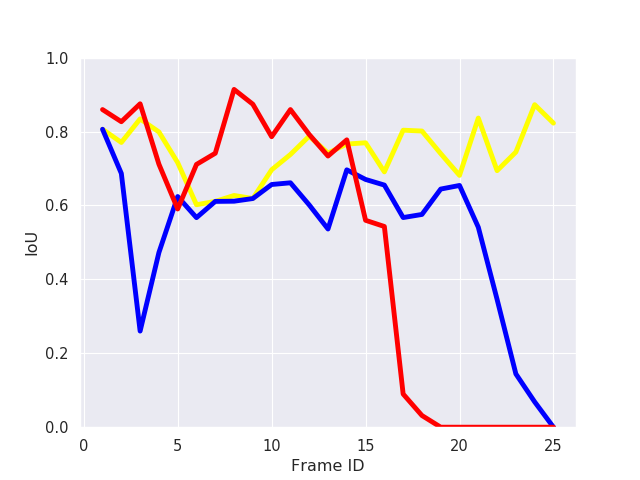}
	}
	\subfigure{
		\includegraphics[width=0.47\linewidth, trim={5 0 10 0}, clip]{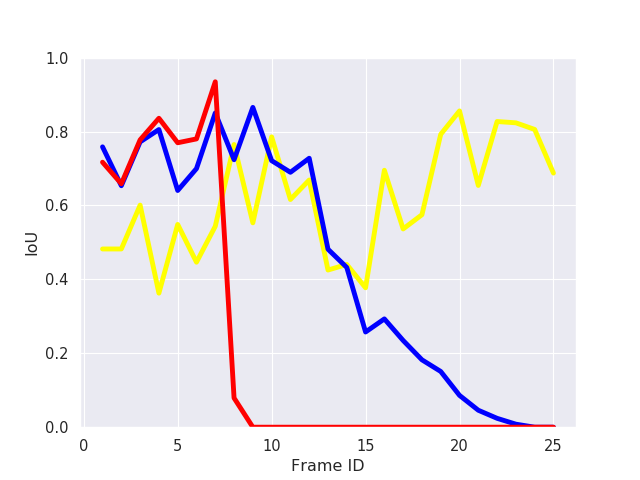}
	}
	\subfigure{
		\includegraphics[width=0.47\linewidth, trim={5 0 10 0}, clip]{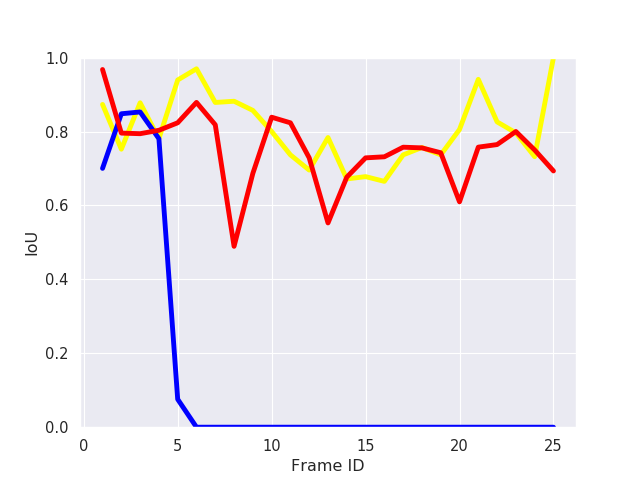}
    }
	\caption{Intersection over Union (IoU) of MOSSE~\cite{mosse} (blue line), of DAT~\cite{nips} (red line), and of our approach (yellow line), with respect to time over 4 test videos. The IoU drops to 0 when a method loses the tracking. MOSSE loses the shadow tracking on every video, the more computationally expensive DAT loses it two out of four times, and our approach does not lose the tracking.}
	\label{fig:IoUframe}
\end{figure}

\begin{figure*}[t!]
	\centering
	\subfigure{
		\includegraphics[width=0.23\linewidth]{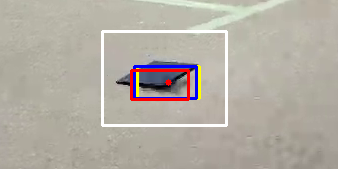}
	}
	\subfigure{
		\includegraphics[width=0.23\linewidth]{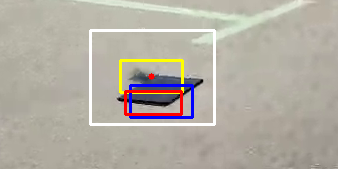}
	}
	\subfigure{
		\includegraphics[width=0.23\linewidth]{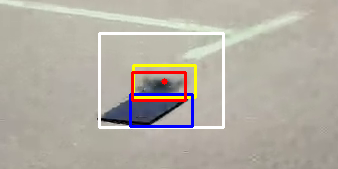}
	}
	\subfigure{
	\includegraphics[width=0.23\linewidth]{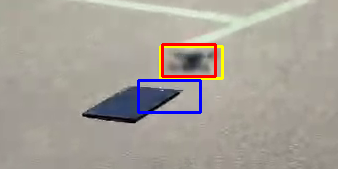}
	}
	\subfigure{
		\includegraphics[width=0.23\linewidth]{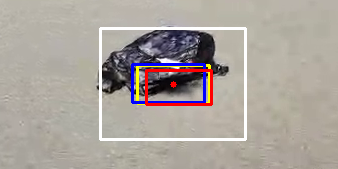}
	}
	\subfigure{
		\includegraphics[width=0.23\linewidth]{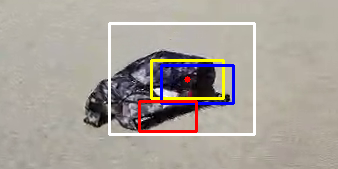}
	}
	\subfigure{
		\includegraphics[width=0.23\linewidth]{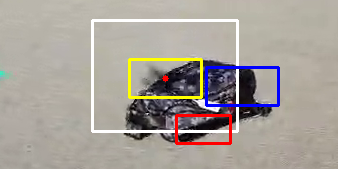}
	}
	\subfigure{
	\includegraphics[width=0.23\linewidth]{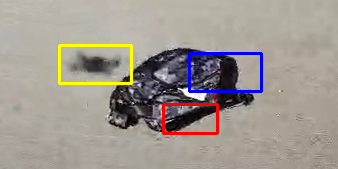}
	}
	\subfigure{
		\includegraphics[width=0.23\linewidth]{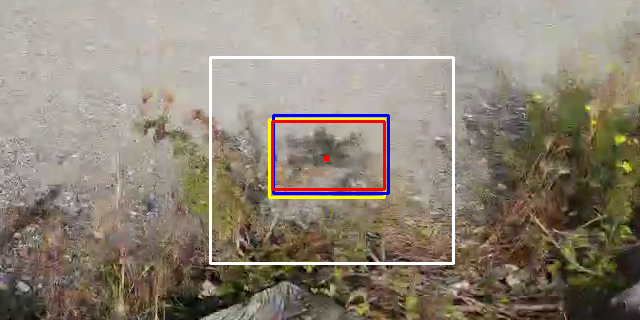}
	}
	\subfigure{
		\includegraphics[width=0.23\linewidth]{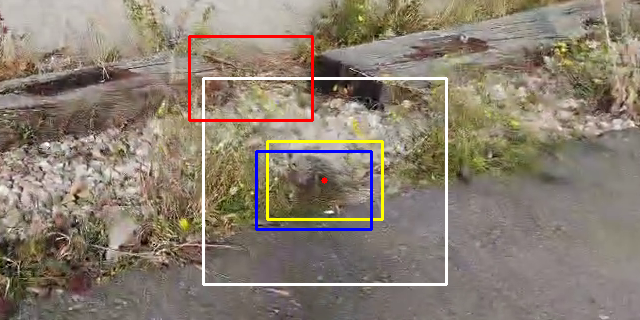}
	}
	\subfigure{
		\includegraphics[width=0.23\linewidth]{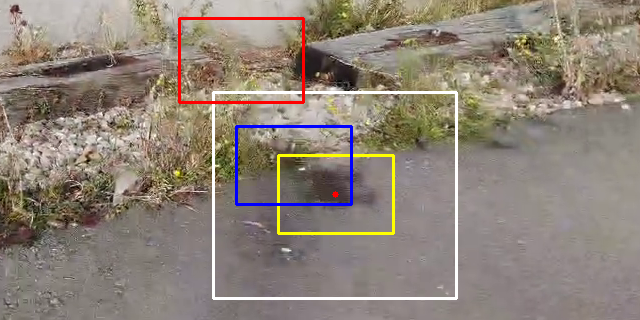}
	}
	\subfigure{
	\includegraphics[width=0.23\linewidth]{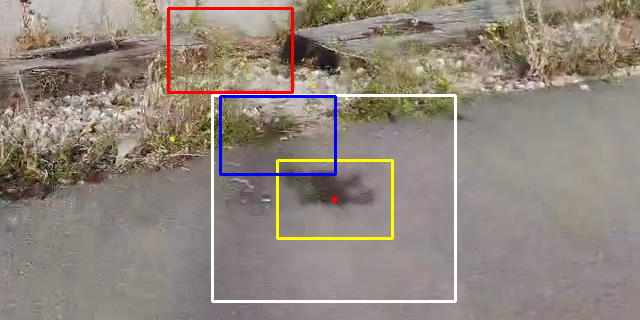}
	}
	\subfigure{
		\includegraphics[width=0.23\linewidth]{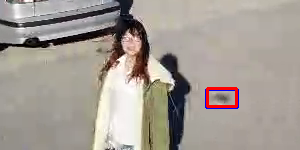}
	}
	\subfigure{
		\includegraphics[width=0.23\linewidth]{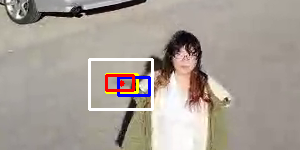}
	}
	\subfigure{
		\includegraphics[width=0.23\linewidth]{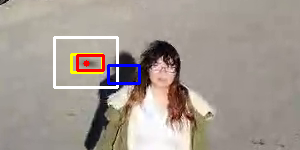}
	}
	\subfigure{
	\includegraphics[width=0.23\linewidth]{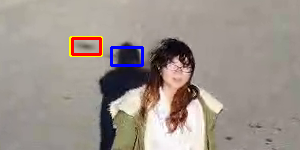}
	}
	\caption{Drone shadow tracking results on the 4 videos with difficult tracking scenes. Each row contains critical frames from one of the 4 videos (with time increasing from left to right). The blue box is the kernel-based tracking result of MOSSE~\cite{mosse}, the red one is the deep learning based tracking result of DAT~\cite{nips} and the yellow box is ours. The white box delimits $S'$.} 
	\label{fig:results}
\end{figure*}

Shadow fusion tracking replaces regular tracking when the failure of the latter is expected. Since kernel-based tracking assumes some spatial contiguity, it fails to recover the tracked shadow after that shadow exits the difficult area where it was lost. This loss is visible in Fig.~\ref{fig:IoUframe} and in the last two columns of Fig.~\ref{fig:results}. 

During shadow fusion tracking, we extend the search area over which correlation analysis is carried and we freeze the kernel update. However, the bigger the search region is, the more noisy and the less reliable the tracking results are. For that reason, tracking algorithms assume spatial contiguity to limit the search area, and this is especially important with objects that have as little consistent features as a drone's shadow. We increase the dimensions of the search region $S$ by a third, chosen heuristically, yielding an extended area
\begin{equation}
    area(S')=\left (\frac{4}{3}\right )^2 \times area(S),
\end{equation}
for the new search region $S'$. We use the same method described in Sec.~\ref{sec:track} for computing the correlation map. However, we set $H_i = H_j$, where $H_j$ is the filter from the last correctly-tracked frame, to avoid inaccurate kernel updates that are likely when failure is likely.

The correlation map $G_i$ is fused with the additional shadow detection map $M_i$ through pixel-wise multiplication.
\begin{equation}
    c_i = \textrm{arg} \max_{c_i\in S'_i} (G_i \odot M_i)
\end{equation}
The peak value of the heat map indicates the center location $c_i$ of the tracking bounding box in the current frame $i$. 

When the drone shadow is lost over a difficult area, three factors contribute to our improved tracking. First, the shadow map contributes to the final heat map, by eliminating all non-shadow locations. Second, the extended search region allows shadow recovery. And third, the use of the last kernel computed before failure was predicted, instead of an incorrectly updated one during likely failure cases, improves the accuracy of the correlation results. 

\section{Evaluation Results}\label{sec:eval}
\subsection{Dataset}\label{sec:dataset}
To evaluate our proposed algorithm, we collect a dataset consisting of 8 drone videos. The drone used for the capture is the DJI Mavic Air. Each video has a $1280 \times 720$ resolution, and a duration of \unit[4]{s} to \unit[10]{s}.

The videos are captured outside, at varying times of the day and with varying degrees of difficulty in terms of tracking the drone's shadow. The video dataset is composed of 4 videos where the scenario is not too complex for the kernel-based tracking, and of 4 videos where the difficulty is higher. In the first set of videos, the drone's shadow is projected on the flat ground, on a green paper bag, on a newspaper and in the fourth video on top of the shadow of a street lamp. In the second set of videos, the drone's shadow is projected on a black laptop, on a black backpack, on grass and pebbles, and lastly on a scene with a moving person. The drone's shadow is prone to getting lost by the tracking on top of the black or very dark objects, on top of the highly textured multi-depth grass and on the moving shadow of the person. 

\subsection{Results}\label{sec:results}
Video annotation is carried out using the VIA image annotator~\cite{dutta2016via} to obtain ground-truth bounding boxes per frame. We annotate 25 frames around the full tracking failure cases and present the average Intersection over Union (IoU) results in Fig.~\ref{fig:IoU} for MOSSE~\cite{mosse}, DAT~\cite{nips}, and our method. IoU results over time for the second set of test videos are also shown in Fig.~\ref{fig:IoUframe}, for each of the three methods.
The state-of-the-art deep learning tracking outperforms MOSSE~\cite{mosse}, but it does so at a computational cost. While the correlation-based tracking can be applied in real time with no specialized hardware, DAT required on average \unit[4.81]{s} per frame, when run on a Titan X GPU. It nonetheless loses the shadow in half of the scenarios.

The kernel-based tracking successfully tracks the drone's shadow in the first set of test videos with simple scenarios. As our results are similar to those of~\cite{mosse} on this set, we do not analyze the corresponding frames. However, on the second set of test videos, MOSSE loses the shadow and fails to recover for the rest of each video, but our tracking does not lose the shadow. Result frames are shown in Fig.~\ref{fig:results} for the 4 videos. The blue box is the result of MOSSE, the red box is the state-of-the-art deep learning tracking (DAT)~\cite{nips}, and the yellow box is ours (videos in supplementary material).

\section{Conclusion} \label{sec:ccl}
We present a shadow tracking approach that leverages the fact that the tracked object has certain physical characteristics, because it is a shadow. Shadow detection is used to predict the failure and improve the performance of correlation-based tracking. The improvement is due to the increased robustness of the heat map fused with shadow detection, the extended search area allowing post-failure recovery, and the freezing of kernel updates during probable failure.

We test our approach on a dataset captured with varying degrees of tracking difficulty, and compare with state-of-the-art deep learning and kernel-based tracking. One application of drone shadow tracking is to remove that shadow for aesthetic purposes. The proposed technique can, however, be leveraged when tracking any shadow or objects with measurable physical properties.

\newpage
\bibliographystyle{IEEEbib.bst}
\bibliography{refs}

\begin{thebibliography}{10}

\bibitem{mosse}
D.~S. Bolme, J.~R. Beveridge, B.~A. Draper, and Y.~M. Lui,
\newblock ``Visual object tracking using adaptive correlation filters,''
\newblock in {\em Proc. IEEE Computer Vision and Pattern Recognition (CVPR)},
  2010, pp. 2544--2550.

\bibitem{inpaintingyu2018}
J.~Yu, Z.~Lin, J.~Yang, X.~Shen, X.~Lu, and T.~S. Huang,
\newblock ``Generative image inpainting with contextual attention,''
\newblock {\em Proc. IEEE Computer Vision and Pattern Recognition (CVPR)},
  2018.

\bibitem{survey}
M.~P. Dhenuka, K.~J. Udesang, and D.~V. Hemant,
\newblock ``Multiple object detection and tracking: A survey,''
\newblock {\em International Journal for Research in Applied Science \&
  Engineering Technology (IJRASET)}, vol. 6, no. 2, 2018.

\bibitem{categories2}
H.~S. Parekh, D.~G. Thakore, and U.~K. Jaliya,
\newblock ``A survey on object detection and tracking methods,''
\newblock {\em International Journal of Innovative Research in Computer and
  Communication Engineering}, vol. 2, no. 2, pp. 2970--2978, 2014.

\bibitem{point_track}
P.~K. Mishra and G.~Saroha,
\newblock ``A study on video surveillance system for object detection and
  tracking,''
\newblock in {\em IEEE International Conference on Computing for Sustainable
  Global Development (INDIACom)}, 2016, pp. 221--226.

\bibitem{kernel_track}
D.~Comaniciu, V.~Ramesh, and P.~Meer,
\newblock ``Kernel-based object tracking,''
\newblock {\em IEEE Transactions on Pattern Analysis and Machine Intelligence},
  vol. 25, no. 5, pp. 564--577, 2003.

\bibitem{silhouette}
J.~Athanesious and P.~Suresh,
\newblock ``Implementation and comparison of kernel and silhouette based object
  tracking,''
\newblock {\em International Journal of Advanced Research in Computer
  Engineering \& Technology}, pp. 1298--1303, 2013.

\bibitem{kcf}
J.~F. Henriques, R.~Caseiro, P.~Martins, and J.~Batista,
\newblock ``High-speed tracking with kernelized correlation filters,''
\newblock {\em IEEE Transactions on Pattern Analysis and Machine Intelligence},
  vol. 37, no. 3, pp. 583--596, 2015.

\bibitem{nips}
S.~Pu, Y.~Song, C.~Ma, H.~Zhang, and M.-H. Yang,
\newblock ``Deep attentive tracking via reciprocative learning,''
\newblock in {\em Advances in Neural Information Processing Systems (NeurIPS)},
  2018, pp. 1935--1945.

\bibitem{survey_removal}
S.~Murali, V.~Govindan, and S.~Kalady,
\newblock ``A survey on shadow removal techniques for single image.,''
\newblock {\em International Journal of Image, Graphics \& Signal Processing},
  vol. 8, no. 12, 2016.

\bibitem{tian}
J.~Tian, L.~Zhu, and Y.~Tang,
\newblock ``Outdoor shadow detection by combining tricolor attenuation and
  intensity,''
\newblock {\em EURASIP Journal on Advances in Signal Processing}, vol. 2012,
  no. 1, pp. 116, 2012.

\bibitem{text}
V.~Shah and V.~Gandhi,
\newblock ``An iterative approach for shadow removal in document images,''
\newblock in {\em Proc. IEEE International Conference on Acoustics, Speech and
  Signal Processing (ICASSP)}, 2018, pp. 1892--1896.

\bibitem{nir}
N.~Salamati, A.~Germain, and S.~S{\"u}sstrunk,
\newblock ``Removing shadows from images using color and near-infrared,''
\newblock in {\em Proc. IEEE International Conference on Image Processing
  (ICIP)}, 2011.

\bibitem{nirour}
D.~R{\"u}fenacht, C.~Fredembach, and S.~S{\"u}sstrunk,
\newblock ``Automatic and accurate shadow detection using near-infrared
  information,''
\newblock {\em IEEE Transactions on Pattern Analysis and Machine Intelligence},
  vol. 36, no. 8, pp. 1672--1678, 2014.

\bibitem{refle}
S.~K. Yarlagadda and F.~Zhu,
\newblock ``A reflectance based method for shadow detection and removal,''
\newblock in {\em Proc. IEEE Southwest Symposium on Image Analysis and
  Interpretation (SSIAI)}, 2018, pp. 9--12.

\bibitem{ma}
L.-Q. Ma, J.~Wang, E.~Shechtman, K.~Sunkavalli, and S.-M. Hu,
\newblock ``Appearance harmonization for single image shadow removal,''
\newblock in {\em Computer Graphics Forum}. Wiley Online Library, 2016,
  vol.~35, pp. 189--197.

\bibitem{guo}
R.~Guo, Q.~Dai, and D.~Hoiem,
\newblock ``Single-image shadow detection and removal using paired regions,''
\newblock in {\em Proc. IEEE Computer Vision and Pattern Recognition (CVPR)},
  2011, pp. 2033--2040.

\bibitem{shluo}
S.~Luo, H.~Li, and H.~Shen,
\newblock ``Shadow removal based on clustering correction of illumination field
  for urban aerial remote sensing images,''
\newblock in {\em Proc. IEEE International Conference on Image Processing
  (ICIP)}, 2017, pp. 485--489.

\bibitem{texture}
F.~Liu and M.~Gleicher,
\newblock ``Texture-consistent shadow removal,''
\newblock in {\em Proc. IEEE European Conference on Computer Vision (ECCV)},
  2008, pp. 437--450.

\bibitem{fourier}
M.~{El Helou}, F.~{D{\"u}mbgen}, R.~{Achanta}, and S.~{S{\"u}sstrunk},
\newblock ``Fourier-domain optimization for image processing,''
\newblock {\em arXiv preprint arXiv:1809.04187}, 2018.

\bibitem{shadow}
D.~Bradley and G.~Roth,
\newblock ``Adaptive thresholding using the integral image,''
\newblock {\em Journal of Graphics Tools}, vol. 12, no. 2, pp. 13--21, 2007.

\bibitem{wellner}
P.~D. Wellner,
\newblock ``Adaptive thresholding for the digitaldesk,''
\newblock {\em Xerox, EPC1993-110}, pp. 1--19, 1993.

\bibitem{dutta2016via}
A.~Dutta, A.~Gupta, and A.~Zissermann,
\newblock ``{VGG} image annotator ({VIA}),''
  www.robots.ox.ac.uk/\texttt{\char`\~}vgg/software/via/, 2016,
\newblock v2.0.5.

\end{thebibliography}

\end{document}